\documentclass[letterpaper, 10 pt, conference]{ieeetran}
\IEEEoverridecommandlockouts                             

\usepackage{graphics}

\usepackage{amsmath}
\usepackage{amssymb}  
\usepackage{siunitx}
\usepackage{mathtools}
\usepackage{booktabs}
\usepackage[ruled]{algorithm2e}
\usepackage{subcaption}
\usepackage[bottom=57pt,top=54pt, left=54pt, right=54pt]{geometry}   

\usepackage{adjustbox}
\usepackage{pbox}
\usepackage{dblfloatfix}
\usepackage{graphicx}
\usepackage[dvipsnames]{xcolor}
\usepackage{hyperref}
\usepackage{caption}
\usepackage{subcaption}
\usepackage[nospace]{cite}

\usepackage{array,multirow,graphicx}
\usepackage{makecell}
\usepackage[normalem]{ulem}

\newcommand{\etal}{\textit{et~al.~}}

\begin{document}
\newgeometry{top=72pt, bottom=57pt, left=54pt, right=54pt}

\title{3D VSG: Long-term Semantic Scene Change Prediction through 3D Variable Scene Graphs}

\author{Samuel Looper$^1$, Javier Rodriguez-Puigvert$^2$, Roland Siegwart$^1$, Cesar Cadena$^1$, and Lukas Schmid$^{1,3}$

\thanks{
This project has received funding from the Microsoft Swiss Joint Research Center, the European Union’s Horizon 2020 research and innovation programme under grant agreement No 101017008 (Harmony), and the Swiss National Science Foundation (SNSF).}%
\thanks{$^1$ Autonomous Systems Lab, ETH Z\"urich, Z\"urich, Switzerland. \newline {\tt\footnotesize \{slooper, rsiegwart, cesarc, schmluk\}@ethz.ch}}%
\thanks{$^2$ Universidad de Zaragoza, Zaragoza, Spain. {\tt\footnotesize jrp@unizar.es}}%
\thanks{$^3$ Massachusetts Institute of Technology, USA. {\tt\footnotesize lschmid@mit.edu}}%
}%

\maketitle

\begin{abstract}
Numerous applications require robots to operate in environments shared with other agents, such as humans or other robots. 
However, such shared scenes are typically subject to different kinds of long-term semantic scene changes.
The ability to model and predict such changes is thus crucial for robot autonomy.
In this work, we formalize the task of \emph{semantic scene variability estimation} and identify three main varieties of semantic scene change: changes in the position of an object, its semantic state, or the composition of a scene as a whole.
To represent this variability, we propose the Variable Scene Graph (VSG), which augments existing 3D Scene Graph (SG) representations with the variability attribute, representing the likelihood of discrete long-term change events. 
We present a novel method, \emph{DeltaVSG}, to estimate the variability of VSGs in a supervised fashion.
We evaluate our method on the 3RScan long-term dataset, showing notable improvements in this novel task over existing approaches.
Our method DeltaVSG achieves an accuracy of 77.1\% and a recall of 72.3\%, often mimicking human intuition about how indoor scenes change over time.
We further show the utility of VSG prediction in the task of active robotic change detection, speeding up task completion by 66.0\% compared to a scene-change-unaware planner.
We make our code available as open-source.

\end{abstract}

\vspace{-4pt}
\section{Introduction}
\label{sec:introduction} 
\vspace{-2pt}

Mobile robotics have the potential to impact numerous applications in healthcare, home robotics, service robotics, or delivery.
These tasks require robots to operate in complex indoor environments that are shared with other agents, such as humans. 
However, shared environments are frequently subject to long-term scene changes, as agents interacting with the scene often cause the position of objects or other semantic scene attributes to change.
This poses a major challenge, as most current methods assume that scenes are static.
Thus, the capacity to capture, model, and predict such changes is essential to enable efficient operation in shared environments.

Current approaches for semantic scene understanding with autonomous robots typically rely on 3D reconstruction of the environment, where each surface element is labeled with the semantic class \cite{semantic_fusion, rosinol_kimera_2020} or object instance \cite{jiang2021indoor,grinvald_volumetric_2019, schmid_panoptic_2022}. 
However, scenes often change in a semantically consistent way, at the level of objects rather than individual surface elements \cite{schmid_panoptic_2022}. 
Humans have an intuition about how our surroundings may change over time. On the one hand, it is unlikely that some objects, such as the fridge and coffee machine, will completely disappear. On the other hand, handheld objects such as mugs can be moved, and typically belong near objects such as tables or drying racks.
Our understanding of how objects change depends significantly on local scene context and relationships between objects. 
Recently, 3D Scene Graphs (SG) \cite{armeni20193d, wald2020learning, rosinol20203d} have emerged as a compact representation that encodes semantic, geometric, and relationship information. 

Scene changes can be classified into two categories: \emph{short-term} dynamics occur within view of the sensor, such as people moving, and \emph{long-term} scene changes, denoting changes beyond the current view of a robot.
This results in notably different change characteristics.
While short-term changes are typically continuous time-series, long-term changes are characterized by discrete and abrupt changes between two observations.
While some SGs model short-term dynamics \cite{rosinol20203d}, scene semantics are typically assumed static.

To model and predict such long-term changes, we propose the concept of Variable Scene Graphs (VSG).
VSGs are an extension of traditional SGs, which additionally model the likelihood of changes occurring for individual objects, which we call \emph{variability}. 
While the presented formalism is general and could also account for short-term changes by placing a high variability on moving objects, we focus on the modeling of discrete long-term changes in this work.
In particular, we formalize \emph{semantic scene variability estimation}, the task of estimating variability for all objects in a scene, to compose a VSG. 
We present DeltaVSG, a novel method to address this task using graph-based learning.
The resulting VSG can predict which parts of a scene are likely to change, allowing robots repeatedly operating in the same environment to harness this predictive power for informed planning based on their previous map.
We demonstrate the utility of VSGs in the task of robotic active change detection.

We make the following contributions:
\begin{itemize}
\vspace{-3pt}
\item We propose 3D Variable Scene Graphs (VSG), a novel formulation to model long-term dynamic scenes and predict scene changes in SGs.
\item We propose DeltaVSG, a method to estimate VSGs, i.e. long-term scene variability, from existing SGs. 
\item We extensively evaluate our method on real world data, showing that accurate VSGs can be generated from standard SGs and demonstrating the potential of VSGs in the task of robotic active change detection. 
We make our code available as open-source\footnote{Released at \url{https://github.com/ethz-asl/3d_vsg}.}.
\end{itemize}

\restoregeometry 
\section{Related Work} 
\label{sec:rel_work}

\subsection{3D Semantic Scene Representations}

Understanding human-made environments is a central topic in robotics.
A first level of semantic understanding can be gained by detecting and classifying object classes or instances in 3D scene reconstructions \cite{han2020occuseg, semantic_fusion, rosinol_kimera_2020, jiang2021indoor, grinvald_volumetric_2019, schmid_panoptic_2022}. 
This has been shown to improve mapping \cite{mccormac2018fusion++} and task planning \cite{yang2018visual} for mobile robots.
Semantic understanding can be further improved by learning object-wise semantic or geometric features \cite{muzahid2020curvenet}, relationships between objects \cite{krishna2017visual}, or scene-specific object attributes \cite{lu2016visual}. 
Several models have been proposed to infer support relationships \cite{silberman2012indoor}, human-scene interactions \cite{hassan2021populating}, and the likelihood of robot co-occurrence \cite{wong2013manipulation}. 
This enhanced semantic understanding has been shown to help in robotic tasks such as object re-localization \cite{druon2020visual}. 

Recently, Scene Graphs (SG) have emerged as a powerful abstract representation of indoor scene semantics.
Initially developed for image understanding \cite{johnson2015image}, they have been extended to 3D and combine various forms of object-level data, as first proposed by Armeni \etal\cite{armeni20193d}.
Wald \etal\cite{wald2020learning} extend SGs with rich semantic attributes, affordances, and relationships.
Rosinol \etal\cite{rosinol20203d} introduce dynamic SGs, providing rich hierarchical abstractions of the SG and accounting for short-term moving agents, such as humans.
Recently, Giuliari \etal\cite{giuliari2022spatial} presented Spatial Commonsense Graphs (SCG), embedding additional nodes from knowledge graphs such as "used for reading", which they call \emph{commonsense concepts}.
Different approaches have been proposed to estimate SGs from images \cite{yang2018graph, lin2020gps} or dense reconstructions \cite{wald2020learning, rosinol20203d}.
Recently, methods to incrementally build SGs have been proposed, such as SGFusion\cite{wu2021scenegraphfusion} or Hydra\cite{hughes2022hydra}, enabling application in online robotics or interaction \cite{tahara2020retargetable}.
Such semantically rich SGs have shown to improve numerous applications, ranging from task planning \cite{agia2022taskography, zhu2021hierarchical}, object retrieval \cite{qiu2020learning, kurenkov2021semantic} or synthetic scene generation \cite{luo2020end, dhamo2021graph}. 
However, these methods assume that scene semantics are static and do not yet account for long-term changes.
Our proposed VSGs thus represent a novel extension of SGs to account for semantic scene variability in long-term changing scenes. 


\subsection{Changes in Scene Semantics}
Semantic scene changes have primarily been studied in the application of change detection.
A first family of approaches focuses on detecting changes in images \cite{daudt2018fully, suzuki2018semantic, park2021changesim}. 
However, these methods rarely reason about semantic changes beyond highlighting differences in image maps. 
As a step towards this, several methods use natural language to classify and caption changes \cite{cheng2020semantic}. 
Ru \etal\cite{ru2020multi} classify semantic changes between two 2D images using an attention-based fusion component, while Kim \etal\cite{kim2021graph} classify scene change by operating on scene graphs. 
While they leverage scene graphs and/or semantics in a similar fashion to our work, they don't capture this knowledge on changes in semantic graphs explicitly to make future predictions. 
These last two methods also don't consider 3D geometry, as they work in a context where geometry can be easily inferred from 2D satellite images. 
Generating semantic-level change annotations is a significant bottleneck for these methods. This can be mitigated using simulated data, as in \cite{li2022exploring}. 

3D spatial information can provide useful geometrical context to semantic data. 
Qiu \etal  use multiple camera views to improve semantic change captioning for 3D scenes \cite{qiu20203d}, while Ku \etal \cite{ku2021shrec} and Li \etal \cite{li2022scene} directly use depth data for image-based change detection. 
Volumetric representations provide even greater context, allowing scene differencing for 3D change detection \cite{fehr2017tsdf, finman_2013_lifelong}.
Such volumetric methods have been extended to utilize semantics for more complete change detection \cite{langer2020robust}, and even during online operation \cite{schmid_panoptic_2022}.
Similar to our work, Liao \etal\cite{liao2021scene} demonstrate how semantic scene graphs can improve change captioning in simple 3D scenes.

However, while change detection involves highlighting semantic differences between a reference and target scene, change prediction is an altogether different task, requiring models to predict the likelihood of future change from a single existing reference scene. 
The ability to predict rather than just react is essential for efficient robot operation in changing environments.
However, this has typically been addressed by predicting change frequencies on individual voxels \cite{frequency_map,wang2020long}. 
While this can already improve planning and localization in changing scenes, such models lack semantic context and consistency provided by SGs.
To this end, Rosinol \etal \cite{rosinol20203d} recently proposed dynamic SGs considering moving agents, whereas Casas \etal \cite{casas2020spagnn} further build a SG to predict the movement of agents in a driving scene. 
However, while this allows for modeling of various short-term dynamic agents like humans and robots in a scene \cite{ravichandran2022hierarchical}, they do not yet consider changes to the scene or objects. 
In contrast, our work focuses on predicting long-term semantic scene changes in both location, semantic attributes, and topology of the individual objects constituting an indoor scene.

\section{3D Variable Scene Graphs}
\label{sec:variability}

\begin{figure*}
    \centering
    \includegraphics[width=\textwidth]{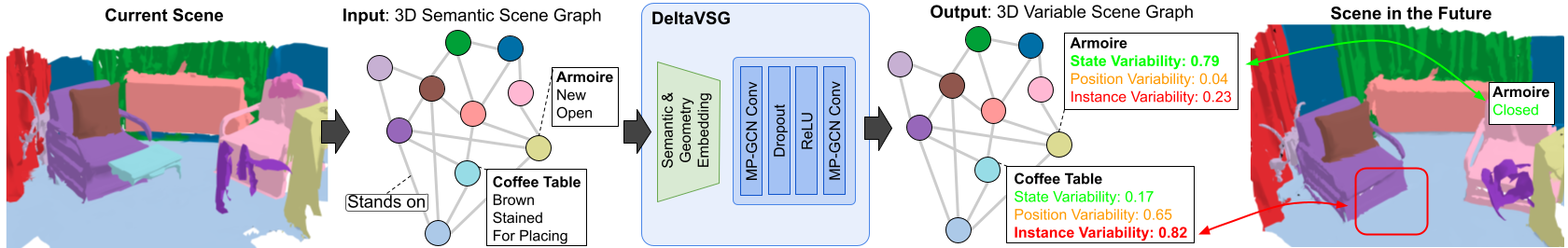}
    \caption{ Given a SG of objects $v_i^t \in \mathcal{V}^t$, attributes $a_i^t \in A^t$, and relations $\mathcal{E}$ of the current scene (left), DeltaVSG estimates a Variable Scene Graph (VSG) by predicting the variability $y_i$ attributes for of each object (center). The VSG can be employed by a robot to predict semantic scene changes for long-term operation (right), or back-propagation of the variability if the future scene is known for training.} 
    \label{fig:overview}
    \vspace{-15pt}
\end{figure*}

Environments shared with other agents inevitably change over time. 
The ability to model and predict such changes is thus essential for long-term autonomy in these environments.
We therefore define \emph{Variable} SGs (VSG) as a representation to capture and predict changing environments, and define the task of \emph{semantic scene variability estimation}.
 
Building on the definition of SGs from Wald \etal \cite{wald2020learning}, we define a VSG $\mathcal{G}$ as a set of vertices $\mathcal{V}$ and a set of edges $\mathcal{E}$, where each entry $v_i \in \mathcal{V}$ represents an object instance in the scene. 
The semantic information of an instance is represented by the object's semantic class $o_i \in O$, and by a set of attributes $a_i \subseteq A$.
The set of possible attributes $A$ includes static properties (e.g. color, rigidity) that are unlikely to change, dynamic properties (e.g. full, open/closed, on/off) that can change as result of interaction, which are called \emph{states}, and the possible interactions an object permits (e.g. sitting, opening) which are called affordances.
Semantic relationships between objects are defined by the set $\mathcal{R}$ and a directional edge mapping $e(v_i, v_j): (\mathcal{V} \times \mathcal{V}) \mapsto \mathcal{R}$. 
Such relationships include support relationships (e.g. standing, lying on), proximity relationships (e.g. next to, in front of), and comparative relationships (e.g. darker than, same shape). 
As scene geometry is an important cue, we additionally include the explicit relative position between two objects as a vector-valued relationship: $r(e_i): \mathcal{E} \mapsto \mathbb{R}^3$. 

To model changes, we define \emph{variability} as the likelihood of a semantic scene change occurring before the next measurement. 
Since scenes typically change in a semantically consistent way, i.e. an entire object is changed rather than an independent piece of space \cite{schmid_panoptic_2022}, the variability $y_i$ can be integrated into the VSG as an additional attribute of each vertex $v_i$.
We identify three major ways a scene can change:
\begin{itemize}
    \item \emph{Position} variability $y_P$ denotes objects moving in space (beyond a minimum threshold), i.e. $r(v_i^t, v_i^{t+1})\geq \epsilon$.
    \item \emph{State} variability $y_S$ models changes in object states, such as a door changing from open to closed, i.e. $a_i^t \neq a_i^{t+1}$.
    \item \emph{Instance} variability $y_I$ denotes topological changes to the graph, i.e. $v_i^t \notin \mathcal{V}^{t+1} \vee v_i^{t+1} \notin \mathcal{V}^t$.
\end{itemize}

While this formulation could also account for transient short-term trajectories by assigning high variability, we focus on long-term changes in this work.
Such long-term changes can result from interactions with the scene outside the current view of the robot, and are thus characterized by abrupt, discrete changes between two measurements.
The task of long-term semantic scene variability estimation can thus be summarized as: estimate $y_i=\{y_P, y_S, y_I\}_i \ \forall n_i \in \mathcal{N}$.
This is a challenging task, as many variations over time are in principle possible, whereas only a single realization occurs.
Lastly, due to our definition of variability, the VSG can directly be employed by a robot for long-term operation, as its attributes are the change prediction.


\section{Approach}
\label{sec:approach}

This section details our method, \emph{DeltaVSG}, to predict long-term scene variability and generate VSGs.
An overview of our approach is shown in Fig.~\ref{fig:overview}.
We discuss scene embeddings in Sec.~\ref{sec:embedding}, followed by the network architecture in Sec.~\ref{sec:learning} and the data processing pipeline in Sec.~\ref{sec:data}.


\subsection{Embedding Semantics and Geometry in Scene Graphs}
\label{sec:embedding} 

Since semantic changes typically occur on the level of objects, we assume the abstract but rich representation of a SG has sufficient information for long-term scene variability prediction.
In addition, since long-term changes typically occur abruptly, we assume a Markovian world and only consider the current state as input.
Our method thus operates on a SG $\mathcal{G}$ as defined in Sec.~\ref{sec:variability} of the current scene to predict the variability $y$.
This allows application of our approach directly on existing SGs, augmenting them to a VSG.
Therefore, we design an embedding function that seeks to embed an input graph $\mathcal{G} = \{\mathcal{V}, \mathcal{E}\}$ containing $N_v=|\mathcal{V}|$ objects and $N_e=|\mathcal{E}|$ edges into a \emph{node embedding matrix} $M_v \in \mathbb{R}^{V_n \times d_v}$, where $d_v$ is the dimension of node embedding vectors, an \emph{edge relationship embedding matrix} $M_R \in \mathbb{R}^{N_e \times N_v}$, and an \emph{edge indices matrix} $M_E \in \mathbb{N}^{N_e \times 2}$, whose rows $e_k = [i, j]\ \forall (i, j) \in \mathcal{E}$.

For node embeddings, semantic information must be compactly represented for neural network inference. 
From the input graph, this is done in a binary encoding vector over a set of discrete features, where each node is assigned a vector $u_i \in [0,1]^{|A|}$ with each $j^\text{th}$ element $u_i^{(j)}$ corresponds to an attribute in the defined taxonomy. 
Without loss of generality, we adopt the classification by Wald \etal \cite{wald2020learning} for this work, which proposes a taxonomy of 92 discrete object semantic attributes and 41 relationship attributes, as well as a hierarchical taxonomy of objects with 527 low-level classes. 
However, any such complex taxonomy results in embedding vectors that are extremely sparse, since objects only belong to a single class and have only a small number of attributes.
We therefore apply principle component analysis (PCA) to reduce the dimensionality of our semantic embedding space. 
We empirically find that using $d_v=120$, the original 619-dimensional binary embedding vectors $u_i$ retain 90\% of their information at up to $\sim5\times$ compression. 

\begin{figure}
    \centering
    \includegraphics[width=\linewidth]{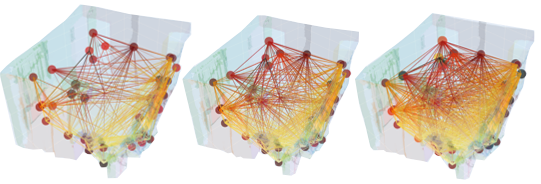}
    \caption{SG relationships at different geometric edge distance thresholds $\tau$, shown for $\tau=$ 1$^{\text{st}}$, 2$^{\text{nd}}$, and 3$^{\text{rd}}$ quartile (left to right).}
    \label{fig:edges}
    \vspace{-15pt}
\end{figure}

Geometry and semantic relationships are combined to create the \emph{edge relationships embedding matrix} $R$. 
For any two objects $v_i, v_j \in \mathcal{V}$, an embedding vector $q_{i,j} \in \mathbb{R}^{|\mathcal{R}|+3}$ is created by concatenating a binary encoding of any semantic relationship between the objects, and the relative position ${}_I\mathbf{r}_{(i, j)} ={}_I\mathbf{r}_j - {}_I\mathbf{r}_i$ where ${}_I\mathbf{r}_i \in \mathbb{R}^3$ represents the position of object $v_i$ in a common reference frame $I$. 
To model spatially local scene context, an edge and associated embedding vector is included for any set of objects $(v_i, v_j)$ where $||{}_I\mathbf{r}_{(i, j)}|| < \tau$ for a distance threshold $\tau$.
This can result in varying levels of connectivity, as illustrated Fig.~\ref{fig:edges}.
Naturally, once the set of object-to-object edges $\mathcal{E}$ and the associated relationship embedding matrix $M_R$ is defined, the \emph{edge indices matrix} $M_E$ can be computed by referring to the index of objects in the node embedding matrix $M_v$. 

\subsection{Learning Scene Graph Variability}
\label{sec:learning}

Given a pair of \emph{current} and \emph{future} scene graphs $(\mathcal{G}^{c},  \mathcal{G}^{f})$, we can formulate long-term scene variability estimation as a supervised learning task.
We first match all objects $V_i^{(c)} \in \mathcal{G}^{c}$ to their counterparts $V_i^{(f)} \in \mathcal{G}^{f}$.
The ground truth variability of $v_i^{(c)}$ can then be computed as $y_{i}  = [y_{P}^{(i)}, y_{S}^{(i)}, y_{I}^{(i)}] \in [0, 1]^3$, as per the definitions of variability in section \ref{sec:variability}.
These labels are aggregated for all objects in a scene $Y = [y_{1}, \dots, y_{N_v}] \in \mathbb{R}^{N_v\times 3}$. 
Lastly, the input graph $\mathcal{G}^{c}$ is embedded to construct a training sample  $X = (M_V^{(c)}, M_E^{(c)}, M_R^{(c)})$, $Y$.
We thus formulate a supervised learning task with the objective of learning a model $\Phi$ that provides an estimated output vector $\tilde{y}_i = \Phi(M_V^{(i)}, M_E^{(i)}, M_R^{(i)})$.

We propose a Graph Neural Network (GNN) architecture to predict the object-wise variability attributes from the resulting embedded SG.
Our proposed architecture to address this learning task, \emph{DeltaVSG}, is based on graph convolutional networks \cite{kipf2016semi}. 
The network utilizes message-passing graph convolution layers (MP-Conv), which computes a latent feature vector $z^{l}_{i}$ for layer $l$ and node $i$ using a standard graph convolution with a nonlinear activation over edge embedding vectors $q_{i,j}$:

\begin{equation}
z^{l+1}_{i} = f_{\theta}(z^{l}_{i}) + \sum_{j \in \text{Neighbors}(i)} z^{l}_{j} \cdot h_{\psi}(q_{i,j}) 
\end{equation}
In this case, $f_{\theta}$ and $h_{\psi}$ are feed-forward neural networks parametrized by weights $\theta$ and $\psi$, respectively.
As with the original image convolution, this offers a form of weight sharing which reduces the overall size of the network needed to learn useful representations. 
As scene changes are oftentimes relatively rare and sparse, we use only two layers of MP-Conv to keep the number of parameters low.
We include a Rectified Linear Unit (ReLU) activation between MP-Conv layers, as well as dropout to prevent overfitting. 
An overview of the architecture is shown in Fig.~\ref{fig:overview}.


\subsection{Data Augmentation and Training Details}
\label{sec:data}
We formulate a dataset $\mathcal{D} = \{(X_1, Y_1), \dots, (X_{N_D}, Y_{N_D})\}$ of samples for the described learning task. 
However, a central limitation in building datasets for semantic scene variability prediction is that collecting a single sample requires scanning and labeling an entire scene at two different times.
In addition, much of the scene may be static.
Therefore, positive samples of scene variability are comparably scarce.
To overcome this, we hypothesize that through long-term object persistence, the temporal direction of long-term scene changes can be neglected, i.e., a cup is equally likely to move from the kitchen to the table as vice-versa.
Thus, we can perform data augmentation by creating unordered and symmetrical pairs from sequences of scans of a scene. 
Given a series of three scans and their corresponding scene graphs $\{\mathcal{G}_1, \mathcal{G}_2, \mathcal{G}_3\}$, we can form training samples from $(\mathcal{G}_1, \mathcal{G}_2)$, $(\mathcal{G}_2, \mathcal{G}_1)$ (symmetrical), $(\mathcal{G}_1, \mathcal{G}_3)$ (unordered) and so on for a total of $n(n-1)$ samples. 

While this significantly increases the number of samples, the data is typically further limited by a high degree of class imbalance. 
In our experiments, only 21\% of nodes had an occurrence of state variability, 17\% for position variability, and 13\% for instance variability. 
We therefore employ a focal loss and perform importance sampling to reduce class imbalance.
Samples with insufficient state information or instance variability are excluded from the position and state variability loss. 
This results in the following element-wise loss function:

\begin{equation}
    \mathnormal{l}_i = -w_c \cdot (1-p_i)^{\gamma}log(p_i)
\end{equation}
where $p_i$ is equal to the probability the model assigns to the label class at element $i$, $\gamma=0.5$ is a hyperparameter controlling the amplification and $w_c$ is the class weight.


\begin{figure*}
    \centering
    \includegraphics[width=\linewidth]{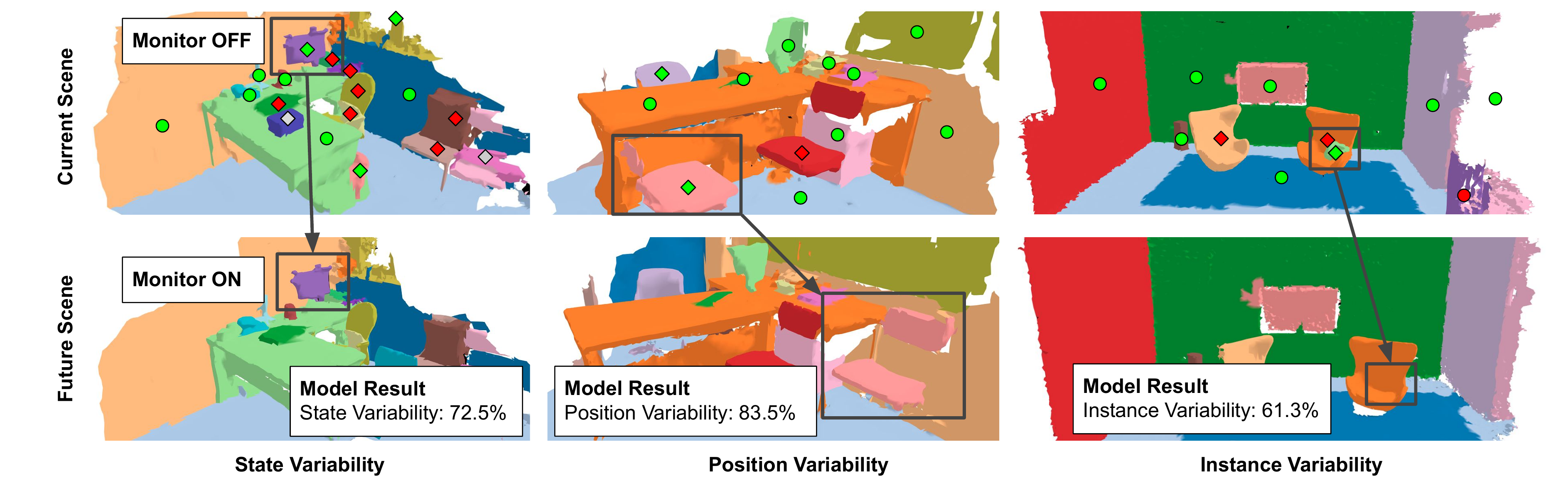}     
    \caption{Qualitative VSG prediction results of DeltaVSG.  We show the current scene (top) overlaid with the vertices $v_i$ of the VSG for different variabilities.  Nodes are shaped if they are predicted to change ($\diamond$) or remain static ($\circ$). The color indicates \textcolor{green}{\textbf{true}} and  \textcolor{red}{\textbf{false}} predictions, as compared to the realized future (bottom). Occurences of positive instance variability are \textcolor{gray}{\textbf{excluded}} from the other variabilities. }
    \label{fig:pred_res}
    \vspace{-5pt}
\end{figure*}

\section{Evaluation} 
\label{sec:evaluation}

In this section, we provide results on the task of \emph{semantic scene variability estimation} introduced in Sec.~\ref{sec:variability}. 
We demonstrate the effectiveness of our approach \emph{DeltaVSG}, compare methods of embedding semantic and geometric information, and show the utility of VSGs for long-term robot operation.

\subsection{Experimental Setup}
\label{sec:experiments}
We train and evaluate our method on the 3DSSG and 3RScan datasets from Wald \etal \cite{wald2019rio, wald2020learning} which include 1482 scans of 478 unique indoor environments changing over time, and their associated scene graphs. 
Repeated measurements were taken on the scale from multiple hours to days, representing the time scale of variability addressed in this experiment. 
We use the ground truth annotations of the dataset to extract the graphs $\mathcal{G}$ and the variability labels Y.
We further perform data augmentation as per Sec.~\ref{sec:data}, generating 3650 effective samples for training. 

To accurately evaluate the quality of the variability prediction, we report individual results on \emph{state}, \emph{position}, and \emph{instance} variability. 
For each case, we present the \emph{accuracy} (Acc.) [\%], i.e. the percentage of correct variation predictions, and the \emph{F1-Score} [\%] of the variation, i.e. the harmonic mean of precision and recall for objects that did change. 
In the tables, the highest number is shown in bold.


\subsection{Variable Scene Graph Prediction Performance}

To evaluate the capacity to estimate accurate VSGs, the scene variability prediction performance of different models is shown in Tab.~\ref{tbl:baseline}. 
We compare our DeltaVSG against two global context baselines, a Multi-Layer Perceptron (MLP) operating on our object class and attribute embeddings, and the LayoutTransformer \cite{gupta2021layouttransformer} using self-attention to model scene context.
We further compare against two graph-based architectures, the GNN of 3DSSG \cite{wald2020learning}, where we replace the PointNet input features with the SG embeddings of \cite{wald2020learning} to operate directly on a SG, and the recent architecture of Spatial Commonsense Graphs (SCG) \cite{giuliari2022spatial} based on graph transformers \cite{shi2020masked}.

We observe that equivalent fully connected architectures, such as MLP, show the lowest performance.
The lack of convolutions may impact the ability to model local context, which significantly decreases recall leading to lower F1 scores.
This is most apparent in position and instance variability.
We further observe that the transformer-based architectures perform marginally worse across all categories than our graph-convolution-based approach.
It is possible that learned attention relationships between object attributes and variability do not generalize as well to this problem. 
We find that our model matches or outperforms the 3DSSG model, which does not explicitly embed relative distances into object relationships. 
DeltaVSG, which explicitly embeds semantic and geometric relationships, thus tends to allow the model to generalize better to less common forms of variability, leading to better overall performance in terms of accuracy and recall. 

Considering all variabilities, DeltaVSG achieves an accuracy of 77.1\% with a recall of 72.3\%.
Across all approaches, models predict a large number of false positives, decreasing precision and thus F1-score, while accuracy and recall remain relatively high. 
We also see that the types of variability with fewer samples and a lower positive sample ratio, namely state and instance variability, show a significant drop in accuracy and F1. 
The overall lower precision reflects the fact that scenes can vary in many equally probable ways due to different possible human interactions.
This reflects the challenging nature of the investigated task, as only one of many plausible future scenarios is realized in the evaluation data.
Nonetheless, DeltaVSG can predict likely scene changes from local semantic and geometric context. 
Qualitative examples in Sec.~\ref{sec:quali} illustrate how this performance translates to a semantically sound understanding of human-driven changes in indoor environments. 
We also show in Sec.~\ref{sec:active_change_detection} that the achieved performance is sufficient to improve a long-term robot operation task.

\begin{table}
\caption{VSG prediction performance of different models.} \vspace{-5pt}
\begin{center}
\begin{tabular}{|l|c|c|c|c|c|c|}
\hline
\multirow{2}{*}{\textbf{Model}}& \multicolumn{2}{c|}{\textbf{State}} & \multicolumn{2}{c|}{\textbf{Position}} & \multicolumn{2}{c|}{\textbf{Instance}}  \\ 
\cline{2-7} 
& \textit{Acc.} & \textit{F1} & \textit{Acc.} & \textit{F1} & \textit{Acc.} & \textit{F1} \\ 
\hline
MLP & 38.9 & 42.0 & 83.3 & 60.0 & 55.3 & 26.5 \\
\hline
LayoutTransformer \cite{gupta2021layouttransformer}& 51.1 & 45.0 & 85.2 & 63.5 & 55.0 & 25.4 \\
\hline
3DSSG \cite{wald2020learning}& 61.5 & 34.2 & \textbf{88.7} & \textbf{68.7} & 63.1 & \textbf{31.6} \\
\hline
SCG \cite{giuliari2022spatial}& 59.7 & 46.3 & 87.0 & 64.9 & 56.4 & 25.8 \\
\hline
DeltaVSG (Ours) & \textbf{62.9} & \textbf{48.9} & 88.1 & 66.6 & \textbf{68.5} & 30.1 \\ 
\hline
\end{tabular}
\label{tbl:baseline}
\end{center}
\vspace{-10pt}
\end{table}


\subsection{Qualitative VSG Prediction Results}
\label{sec:quali}

Qualitative results of the VSG prediction are shown in Fig.~\ref{fig:pred_res}, which illustrate how DeltaVSG predictions often capture human intuition on scene variability. 
The model correctly identifies small handheld objects and frequently used objects such as chairs with a high probability of position or instance variability, as opposed to large pieces of furniture like desks or rugs. 
At a high level, DeltaVSG predictions are often informed by the larger layout of the scene. 
A desk configuration more consistent with working use leads to higher probabilities of a state change for a computer monitor. 
Chairs laid out for a temporary face-to-face meeting are predicted to move back in place. 
While some of these intuitively likely predictions yield false positives when compared to the specific future scene in the training sample, they still represent a consistent understanding of how objects can change in human-made environments.


\subsection{Comparing Embedding Methods}

We present several ablation experiments on the impact of embedding methods for DeltaVSG. 
First, we study the impact of local geometric context in Tab.~\ref{tbl:geom_res} by varying the distance $\tau$ for establishing geometric relationship edges.
Overall, we find a performance improvement when local geometric edges are added to the graph.
However, this effect is lessened when many further away objects are considered, and can even reduce model performance.
This indicates the importance of primarily local scene context, which appears to be sufficient information for semantic scene change prediction.

Second, we investigate the impact of the semantic taxonomy, i.e. the number distinguished semantic classes, in Tab.~\ref{tbl:taxonomies}.
Note that the attributes $A$ and thus the difficulty of the variability estimation task remain unchanged.
We note a tradeoff between performance improvements caused by finer-grained semantic information, and degrading performance due to sparser embedding vectors and data samples.
However, while our initial model is trained on RIO527 \cite{wald2020learning}, generally high performance is maintained also when using taxonomies with a significantly reduced class set. 

Lastly, we discuss the importance of compact embedding in Tab.~\ref{tbl:embedding}.
We note that our final method combining the large RIO object taxonomy and dimensionality reduction leads to the highest performance.
This further hints at the trade-off between compact embeddings in learning tasks with sparse positive samples and detailed semantic information.

\begin{table}
\caption{DeltaVSG performance at different geometrical relationship thresholds $\tau$ as a percentile of the training distribution.} \vspace{-8pt}
\begin{center}
\begin{tabular}{|l|c|c|c|c|c|c|c|}
\hline
\multirow{2}{*}{\textbf{Percentile}} & \multirow{2}{*}{$\tau$ [m]} & \multicolumn{2}{c|}{\textbf{State}} & \multicolumn{2}{c|}{\textbf{Position}} & \multicolumn{2}{c|}{\textbf{Instance}}  \\ 
\cline{3-8} 
& & \textit{Acc.} & \textit{F1} & \textit{Acc.} & \textit{F1} & \textit{Acc.} & \textit{F1} \\ 
\hline
0$^{th}$ & - & 52.1 & 42.4 & 84.2 & 53.6 & 58.0 & 37.1 \\
\hline
25$^{th}$ &  1.67 & 54.5 & 41.5 & 87.1 & 66.5 & 64.3 & \textbf{37.5} \\
\hline
50$^{th}$ & 2.61 & \textbf{66.6} & 35.1 & 87.2 & 63.7 & \textbf{68.7} & 36.7 \\ 
\hline
75$^{th}$ &  3.70 & 62.9 & \textbf{48.9} & \textbf{88.1} & \textbf{66.6} & 68.5 & 30.1 \\ 
\hline
100$^{th}$ & 15.5 & 53.9 & 35.0 & 86.3 & 59.2 & 58.3 & 24.6 \\ 
\hline
\end{tabular}
\label{tbl:geom_res}
\end{center}
\end{table}

\begin{table}
\caption{DeltaVSG performance on different class taxonomies.}\vspace{-8pt}
\begin{center}
\begin{tabular}{|l|c|c|c|c|c|c|}
\hline
\multirow{2}{*}{\textbf{Taxonomy}}& \multicolumn{2}{c|}{\textbf{State}} & \multicolumn{2}{c|}{\textbf{Position}} & \multicolumn{2}{c|}{\textbf{Instance}}  \\ 
\cline{2-7} 
& \textit{Acc.} & \textit{F1} & \textit{Acc.} & \textit{F1} & \textit{Acc.} & \textit{F1} \\ 
\hline
RIO527 \cite{wald2020learning} & 62.9 & \textbf{48.9} & \textbf{88.1} & \textbf{66.6} & \textbf{68.5} & 30.1 \\ 
\hline
NYU40 \cite{silberman2012indoor} & 42.3 & 31.0 & 84.6 & 57.0 & 57.5 & 27.3 \\
\hline
RIO27 \cite{wald2019rio} & 58.6 & 36.1 & 85.8 & 58.7 & 63.1 & \textbf{31.0} \\
\hline
Eigen13 \cite{couprie2013indoor} &\textbf{64.1} & 40.8 & 86.9 & 59.4 & 59.4 & 27.1 \\
\hline
\end{tabular}
\label{tbl:taxonomies}
\end{center}
\vspace{-10pt}
\end{table}

\begin{table}
\caption{DeltaVSG performance with different semantic object class embedding sizes $d_v$.}\vspace{-8pt}
\begin{center}
\begin{tabular}{|l|c|c|c|c|c|c|c|}
\hline
\multirow{2}{*}{\textbf{Embedding}} & \multirow{2}{*}{$d_v$} & \multicolumn{2}{c|}{\textbf{State}} & \multicolumn{2}{c|}{\textbf{Position}} & \multicolumn{2}{c|}{\textbf{Instance}}  \\ 
\cline{3-8} 
& & \textit{Acc.} & \textit{F1} & \textit{Acc.} & \textit{F1} & \textit{Acc.} & \textit{F1} \\ 
\hline
Binary & 619 & \textbf{81.4} & 0.0 & 71.3 & 39.1 & \textbf{87.8} & 0.0 \\
\hline
PCA & 120 & 62.9 & \textbf{48.9} & \textbf{88.1} & \textbf{66.6} & 68.5 & \textbf{30.1} \\
\hline
PCA & 48 & 58.8 & 35.5 & 85.6 & 58.6 & 70.2 & 28.1 \\
\hline
\end{tabular}
\label{tbl:embedding}
\end{center}
\end{table}

\subsection{Application to Active Change Detection}
\label{sec:active_change_detection}

Lastly, we demonstrate the utility of VSGs in the representative application of active change detection.
In this task, a robot can move from object to object based on the map from the previous time, and at each location measure whether the object has changed.
The goal is to identify $n$ changes in the scene.

As a baseline, we employ an optimal \emph{Coverage} path by casting all objects of the previous map into a Traveling Salesman Problem (TSP) \cite{junger1995traveling}.
In comparison, we show the utility of VSGs using a variability-aware method.
\emph{VSG-Planner} computes the VSG from the previous map using our DeltaVSG network, and then solves the TSP for the $n+3$ objects with the highest variability probabilities.
If less than $n$ changes are observed within that path, the robot resorts to \emph{Coverage} for the remaining objects.

\begin{figure}
    \centering
    \includegraphics[width=\linewidth]{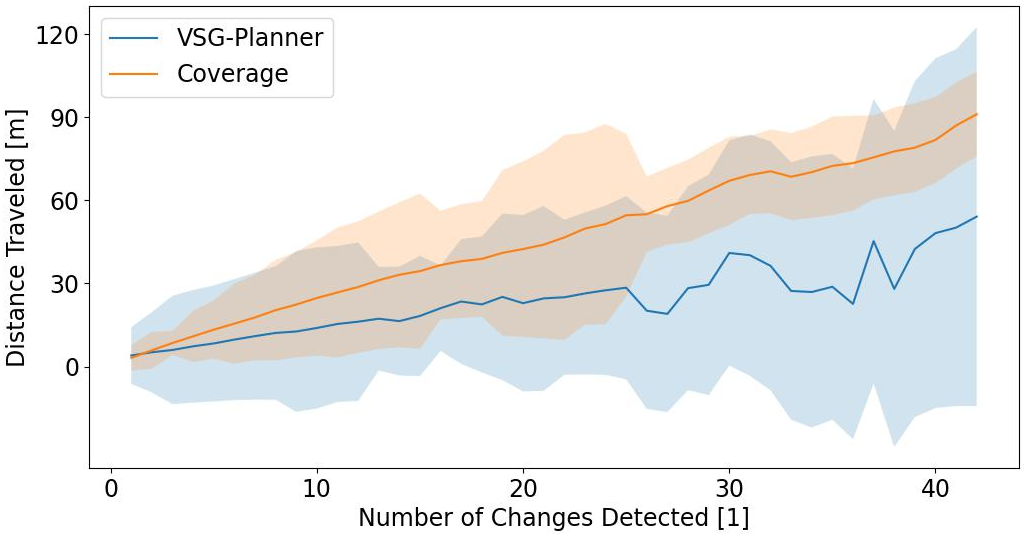}
    \caption{Mean and standard deviation of distance traveled until $n$ changed objects are detected in the active change detection task.}
    \label{fig:task}
    \vspace{-10pt}
\end{figure}

The performance of both methods is shown in Fig.~\ref{fig:task}.
Overall, we observe that our DeltaVSG model can be effectively used to improve task efficiency, especially when a larger number of changes must be detected. 
Again, we observe the challenge in long-term scene change prediction of different possible futures existing, reflected in the high variances.
Nonetheless, VSG-Planner is able to identify a shorter path in 68.7\% of cases, speeding up change detection by 66.0\% on average.

As such, we believe a robot performing inspections, collecting data, or performing other continuous tasks in human environments could leverage VSGs to assign uncertainty to the prior scene graph or environment map.

\section{Conclusions}
\label{sec:conclusion}
In this work, we addressed the challenge of modeling semantic scene changes for robots continuously operating in shared environments.
We thus formalized the problem of \emph{semantic scene variability estimation}, identifying three major types of long-term semantic scene change, and formulate it as a supervised learning task. 
We presented 3D Variable Scene Graphs (VSG) as a natural and compact representation of scene variability.
Lastly, we developed \emph{DeltaVSG}, a novel approach operating on existing SG representations and combining explicit semantic and geometric embeddings with a GNN architecture to estimate the resulting VSG.
We show in thorough experimental evaluation that our approach is able to capture intuitive scene variability predictions, achieving accuracy of 77.1\% and recall of 72.3\% on this challenging task.
We demonstrate the utility of VSGs in the task of long-term change detection, speeding up task completion by an average of 66.0\% compared to variability-unaware methods.

While we presented a first approach to estimate VSGs, there are numerous directions for future research in this task of semantic scene variability estimation.
While we primarily focused on predicting change events, these can be extended to more detailed predictions, such as predicting the exact future state, where an object is likely to move, or also accounting for other topological changes such as where and which objects are likely to appear.
Lastly, we hope to encourage more research in applying semantic change predictions and VSGs to different long-term robot autonomy tasks.





{\small
\bibliographystyle{IEEEtran}
\bibliography{references}
}

\end{document}